\documentclass{article} 
\usepackage{iclr2027_conference,times}

\usepackage{amsmath,amsfonts,bm}

\def\eqref#1{equation~\ref{#1}}

\def\1{\bm{1}}

\DeclareMathAlphabet{\mathsfit}{\encodingdefault}{\sfdefault}{m}{sl}
\SetMathAlphabet{\mathsfit}{bold}{\encodingdefault}{\sfdefault}{bx}{n}

\usepackage{hyperref}
\usepackage{url}
\hypersetup{hidelinks}
\usepackage{graphicx} 
\usepackage{tikz}
\usetikzlibrary{
    positioning,
    arrows.meta,
    calc,
    shapes.geometric
}
\usepackage{booktabs}
\usepackage{svg}
\usepackage{subcaption}
\usepackage{placeins}
\title{Outcome-Balanced PPO and Policy-Induced Flow Rectification for Flow-Based VLA Policies}

\author{Yuhan Wang$^{1}$, Ruiyang Wang$^{1}$, Jiaguan Zhu$^{2}$ \\
$^{1}$Tsinghua University \\
$^{2}$The Hong Kong University of Science and Technology
}

\iclrfinalcopy
\renewcommand{\headrulewidth}{0pt}
\begin{document}

\maketitle
\fancyhead{}
\begin{abstract}
Reinforcement learning (RL) enables post-training of flow-based vision-language-action (VLA) policies by optimizing expected task return through environment interaction. However, this optimization raises two challenges: policy updates can alter the velocity-field geometry underlying the score approximation used for stochastic transitions, and sparse multi-task rollouts can yield uneven learning signals across tasks and advantage signs. We introduce an RL framework that combines Outcome-Balanced Proximal Policy Optimization (PPO) with policy-induced flow rectification. Outcome-Balanced PPO reweights positive advantages using task success-rate estimates and rescales positive and negative advantages while preserving their signs. Policy-induced rectification periodically fits a linear-bridge flow to ODE endpoint pairs generated by the current policy, refreshing the representation used in subsequent stochastic optimization. We evaluate the framework on simulated manipulation benchmarks and additionally study a component combination in a Real2Sim2Real task, examining task performance, optimization behavior, and flow consistency. The proposed framework improves success rates over the evaluated baselines across the simulated benchmarks.
\end{abstract}

\section{Introduction}

Flow-based vision-language-action (VLA) models, including $\pi_0$ and $\pi_{0.5}$, generate continuous action chunks using learned flows conditioned on visual observations and language instructions~\citep{black2024pi0,black2025pi05}. Their expressive action distributions and ODE-based samplers support robotic manipulation and provide a foundation for online reinforcement learning (RL), in which task rewards can refine behavior beyond that represented in the demonstrations. Recent empirical evidence supports PPO as an effective approach to online RL for VLA policies, particularly under sparse rewards~\citep{liu2025can}.

Flow-GRPO enables tractable online optimization of flow policies by constructing an SDE with the same marginals as the probability-flow ODE and deriving transition likelihoods under a linear-path assumption~\citep{flow_grpo}. In this construction, the score approximation used in stochastic transitions depends on the geometry of the underlying flow. Task-driven RL updates alter the velocity field, potentially reducing the accuracy of the original approximation for the policy-induced intermediate distributions.

Sparse multi-task training can also produce uneven policy updates. Tasks with low success rates may provide limited positive learning signals, while positive and negative advantages may differ substantially in magnitude. These observations motivate two questions: (1) how can stochastic policy optimization remain reliable as the velocity-field geometry changes? and (2) how can learning signals be allocated more deliberately across tasks and advantage signs?

We address these questions with two complementary mechanisms. \emph{Outcome-Balanced PPO} uses an exponential moving average of each task's success rate to reweight positive advantages, followed by sign-preserving scaling of positive and negative advantages. It modifies the allocation and scale of the PPO learning signals without changing trajectory sampling or the clipping rule. \emph{Policy-Induced Flow Rectification} periodically pauses PPO, collects deterministic ODE endpoint pairs from the current policy, and fits a linear-bridge flow to the resulting coupling. This separate reconstruction stage updates the representation used in subsequent SDE-based optimization while allowing task-driven adaptation between rectification stages.

Our contributions are summarized as follows:
\begin{itemize}
    \item We study two issues in RL fine-tuning of flow-based VLA policies: unequal learning-signal scales across tasks and advantage signs, and mismatch between an RL-adapted velocity field and the linear-path geometry used by Flow-SDE score estimation.

    \item We propose Outcome-Balanced PPO, which weights positive advantages using task success-rate estimates and separately rescales positive and negative advantages while preserving their signs.

    \item We propose periodic policy-induced flow rectification, a separate phase that pauses PPO and reconstructs a linear-bridge velocity field from deterministic ODE endpoint pairs generated by the current policy.

    \item We evaluate the combined framework on simulated manipulation benchmarks and assess a combination of its components in a Real2Sim2Real task.
\end{itemize}

\begin{figure}
    \centering
    \includegraphics[width=0.9\linewidth]{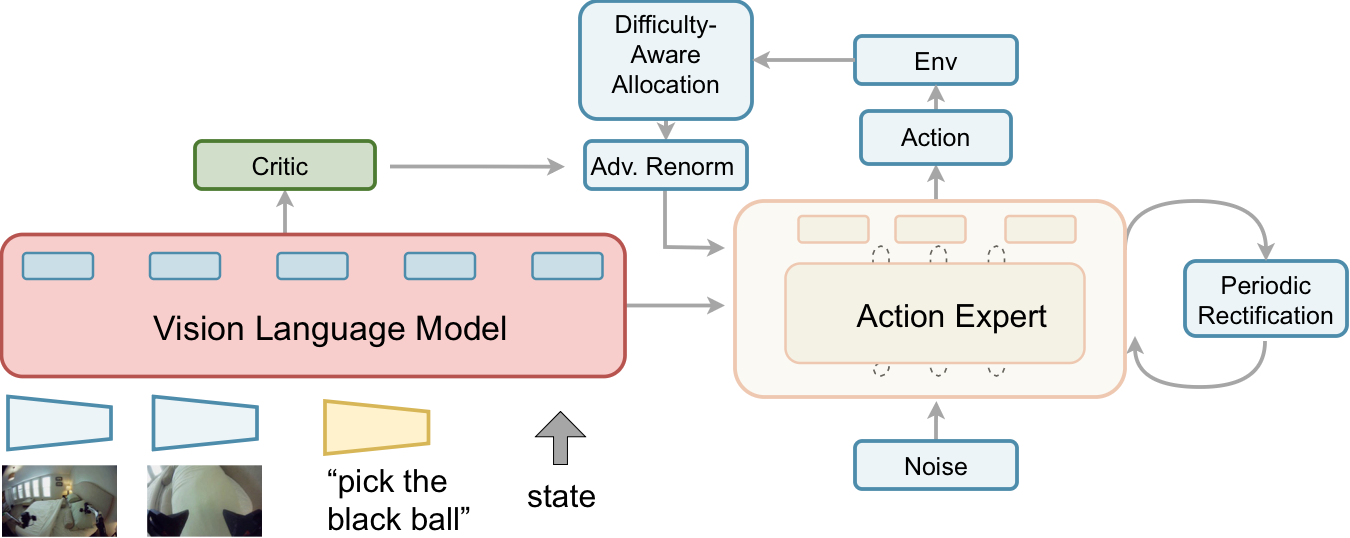}
    \caption{
    Overview of the proposed RL post-training framework for $\pi_{0.5}$.
    The vision-language module is frozen and provides a shared multimodal
    representation to both the critic and the trainable flow-based
    action expert. 
    The action expert generates action chunks through the flow ODE/SDE and is
    optimized with Outcome-Balanced PPO. When rectification is triggered, PPO updates
    are paused and the flow is reconstructed from endpoint pairs generated by
    the current policy, after which PPO training resumes.
    }
    \label{fig:method-overview}
\end{figure}

\section{Related Work}
\paragraph{Reinforcement Learning for Flow-Based Policies.}
Flow matching provides an expressive representation of continuous policies
for robotic control and VLA models. Flow-GRPO~\citep{flow_grpo} constructs
an SDE with the same marginals as the flow ODE, enabling stochastic
exploration and policy-gradient optimization. $\pi_{RL}$~\citep{pi_rl}
extends this ODE--SDE formulation to flow-based VLA policies.
ReinFlow~\citep{reinflow} introduces learnable noise into deterministic
flow trajectories and formulates the policy as a discrete-time stochastic
process with tractable transition likelihoods. Flow Policy Optimization
(FPO)~\citep{fpo} instead derives policy-gradient objectives from
conditional flow matching, avoiding explicit evaluation of the underlying
flow density. Score-Flow~\citep{score-flow} decouples the mean and variance
of SDE transitions, providing separate control over deterministic drift
and stochastic dispersion. The score constructions used in these
ODE--SDE approaches depend on the assumed flow formulation. Our work
examines how RL updates affect the compatibility between the velocity
field and the score approximation used for stochastic optimization.

Other approaches modify the policy parameterization or optimization
objective. RoboFFT~\citep{robofft} uses a forward process and a surrogate
ratio for PPO-style fine-tuning, while PrT-Flow~\citep{prtflow} makes
finite-step flow transitions invertible to enable density evaluation for
actor--critic RL. ReFPO~\citep{refpo} incorporates reflow regularization
into flow-matching policy gradients. These methods address likelihood
construction, policy parameterization, or flow regularity.

\paragraph{Advantage Reweighting and Asymmetric Policy Optimization.}
Advantage estimates determine the direction and relative strength of
policy updates, motivating methods that explicitly control how these
signals enter the optimization objective. Advantage-Weighted Regression
(AWR)~\citep{awr} and Advantage-Weighted Actor-Critic
(AWAC)~\citep{awac} use exponential advantage weights in regression
or maximum-likelihood policy updates, emphasizing higher-advantage
actions while supporting learning from off-policy data. For flow-based
VLA policies, ARFM~\citep{arfm} adopts advantage-weighted flow matching
for offline post-training. It adaptively adjusts the exponential
weighting strength to balance advantage emphasis and gradient variance.
These approaches use positive regression weights, so lower-advantage
samples receive less relative emphasis rather than contributing
signed policy-gradient updates.

A complementary line of work studies the asymmetric roles of positive
and negative learning signals. Asymmetric
REINFORCE~\citep{asymmetric_reinforce} analyzes how the choice of
baseline changes their relative influence, establishing conditions
under which emphasizing positive rewards benefits off-policy policy
improvement. BAPO~\citep{bapo} identifies imbalanced positive and negative
contributions in off-policy PPO-style training and adaptively adjusts
clipping bounds to regulate their contributions to the policy loss.
A3PO~\citep{a3po} instead introduces asymmetric token-level advantage
shaping for reinforcement learning with verifiable rewards.
EAPO~\citep{eapo} adjusts positive-sample weights according to the ratio
between current and initial policy entropy, using this signal to
regulate exploration and exploitation. These methods address
asymmetry through baseline selection, adaptive clipping, token-level
shaping, or entropy-dependent weighting.

Learning-signal allocation also arises in multi-task reinforcement
learning. PopArt~\citep{popart} uses adaptive normalization to address
differences in return scales that can cause some tasks to dominate
learning. This highlights the distinction between reward-scale
imbalance and differences in task success: normalizing return scales
does not directly allocate additional positive learning signal to
tasks with lower success rates.

Our Outcome-Balanced PPO addresses both task-level allocation and
sign-dependent advantage scales. It first uses task success-rate
exponential moving averages to assign bounded weights to positive
advantages, increasing their relative emphasis on tasks with lower
estimated success. It then separately rescales positive and negative
advantages to a shared mean absolute magnitude while preserving
their signs. The resulting signed advantages are used in the standard
clipped PPO objective. The method therefore differs from positive-weighted
regression,
baseline shifts, and adaptive clipping. Its adaptation signals are
task success rates and batch-level advantage statistics, rather than
policy entropy or token-level credit assignments. The paired scaling
matches group-level advantage
magnitudes; it does not enforce equal sample counts or equal gradient
contributions.

\paragraph{Flow Rectification and Coupling Reconstruction.}

Rectified Flow~\citep{rectified_flow} learns velocity fields along linear
interpolations between source and target samples. Reflow reconstructs the
transport coupling from ODE-generated endpoint pairs and retrains the flow
on this coupling; repeated reconstruction can reduce trajectory curvature
and improve sampling efficiency. Subsequent work has explored progressive
and consistency-based variants of flow rectification~\citep{proreflow}.
ReFPO~\citep{refpo} incorporates a reflow loss, weighted by $\lambda$,
into the FPO objective. This joint objective couples policy optimization
and flow regularization through their shared gradient updates, with the
trade-off controlled by $\lambda$. Our method instead separates PPO
updates from flow reconstruction. It uses ODE endpoint pairs generated by
the current RL policy to define a policy-induced coupling and periodically
reconstructs the velocity field during a PPO-pausing stage. The resulting
representation is then used in subsequent stochastic optimization.

\section{Preliminaries}
\paragraph{Flow Matching for Vision-Language-Action Models.}

Let $c$ denote the observation-language context. Let $\mathbf{X}_1 \sim p_1$ be a sample from a simple prior distribution, typically an isotropic Gaussian, and let $\mathbf{X}_0 \sim p_0(\cdot\mid c)$ be an action chunk drawn from the target conditional distribution. We define a linear interpolation path between $x_1$ and $x_0$ as
\begin{equation}
    x_t=t x_1+(1-t)x_0
\label{equ:linear interpolation}
\end{equation}

The corresponding target velocity is $v_{\mathrm{target}}=x_1-x_0$. The flow-based action expert is trained to approximate this velocity by minimizing the conditional flow-matching loss:
\begin{equation}
    \mathcal{L}(\theta)=\mathbb{E}_{t,x_0 \sim X_0, x_1 \sim X_1}[||v_\theta(x_t,t,c)-v_{target}||^2]
\end{equation}

\paragraph{ODE--SDE Action Generation.}

To enable stochastic exploration during RL post-training, Flow-GRPO~\citep{flow_grpo} introduces an SDE whose marginals match those of the corresponding ODE:
\begin{equation}
    dx_t=\left(v_\theta(x_t,t,c)-\frac{\sigma_t^2}{2}\nabla \log p_t(x_t)\right)dt + \sigma_t d\omega.
\end{equation}
Here, $\sigma_t$ controls the noise magnitude at time $t$. Under the linear-path assumption in Eq.~\eqref{equ:linear interpolation}, the score $\nabla \log p_t(x_t)$ has the following closed-form expression:
\begin{equation}
    \label{eq:linear score}
    \nabla \log p_t(x_t)=-\frac{x_t}{t}-\frac{1-t}{t}v_\theta(x_t,t,c)
\end{equation}

\paragraph{PPO Optimization of Flow Policies.}
Under the ODE--SDE formulation, adjacent states along a flow trajectory define conditional Gaussian transitions. The Euler--Maruyama discretization gives $x_{t_{k-1}}\sim \mathcal{N}(\mu_\theta(x_{t_k},t_k),\sigma_{t_k}^2\Delta_{t_k}\mathbf{I})$, where $\mu_\theta(x_{t_k},t_k)=x_{t_k}+\left[v_\theta(x_{t_k},t_k)+\frac{\sigma_{t_k}^2}{2t_k}(x_{t_k}+(1-t)v_\theta(x_{t_k},t_k))  \right]\Delta t_k$.

The conditional transition likelihoods yield the following factorization of the flow-trajectory likelihood:
$
\pi_{\theta}(\tau\mid c)
=
\prod_{k=1}^{K}
\pi_{\theta}
\left(
x_{t_{k-1}}
\mid
x_{t_k},c
\right),
$
The transition-level likelihood ratio can then be incorporated into the clipped PPO objective:
\begin{equation}
\mathcal{L}_{\mathrm{PPO}}
=
-
\mathbb{E}
\left[
\min
\left(
r_k(\theta)\hat{A}_k,
\operatorname{clip}
\left(
r_k(\theta),1-\epsilon,1+\epsilon
\right)
\hat{A}_k
\right)
\right], 
\quad r_{k}(\theta)
    =
    \frac{
        \pi_{\theta}(x_{t_{k-1}}\mid x_{t_k},c)
    }{
        \pi_{\theta_{\mathrm{old}}}
        (x_{t_{k-1}}\mid x_{t_k},c)
    }
    \label{equ:loss ppo}
\end{equation}
where $\hat{A}_k$ is the generalized advantage estimate, $r_k(\theta)$ is the transition-level importance ratio, and $\epsilon$ is the PPO clipping coefficient. The Gaussian transition model therefore provides tractable likelihoods for flow-based VLA policies, enabling direct optimization of the velocity field with PPO while retaining stochastic exploration.

\section{Methods}

Our framework addresses two complementary aspects of RL post-training for
flow-based policies. Section~\ref{sec:policy-induced-rectification}
describes policy-induced flow rectification, which reconstructs the
velocity field used in subsequent stochastic optimization.
Section~\ref{sec:outcome_balanced_ppo} presents Outcome-Balanced PPO,
which reallocates and rescales advantage signals during policy updates.

\subsection{Policy-Induced Flow Rectification}
\label{sec:policy-induced-rectification}
\begin{figure}
    \centering
    \includegraphics[width=0.99\linewidth]{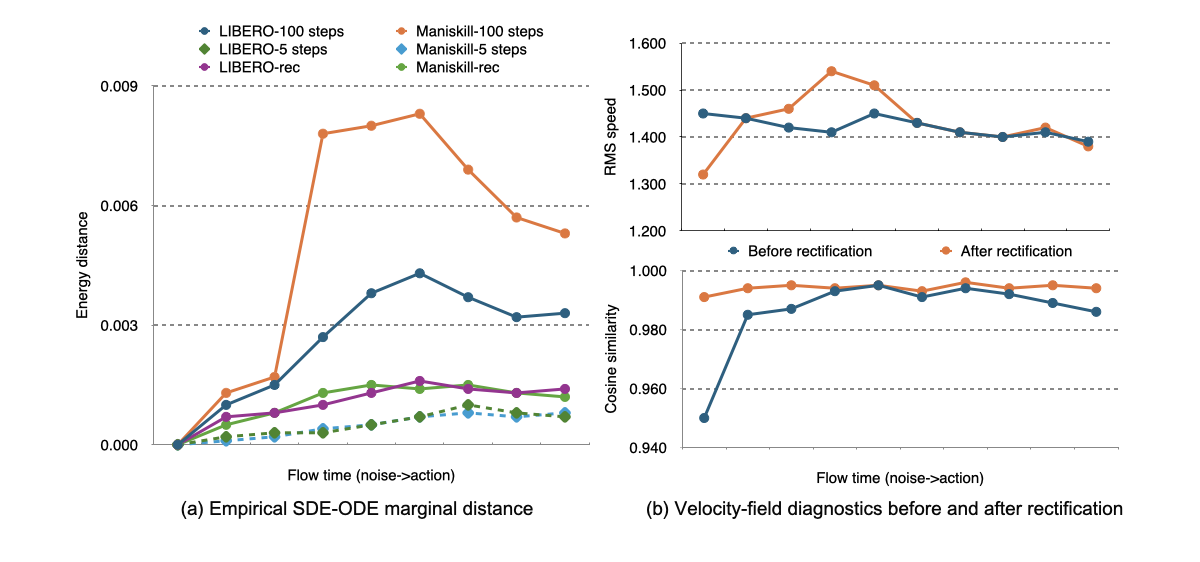}
    \caption{Flow diagnostics. (a) SDE--ODE marginal discrepancy before RL, after RL updates, and after policy-induced rectification. (b) Root-mean-square (RMS) speed and cosine similarity across flow timesteps before and after rectification.}
    \label{fig:flow-analysis}
\end{figure}

Equation~\eqref{eq:linear score} gives a score estimate under the
linear-path assumption used to construct the SDE. PPO updates can alter
the velocity field nonuniformly across flow time, so the updated field
may no longer satisfy the same path assumption. The resulting estimate
may therefore differ from the score of the policy-induced intermediate
marginal. Figure~\ref{fig:flow-analysis}(a) reports the empirical SDE--ODE
marginal discrepancy across flow time for ManiSkill and LIBERO-10,
illustrating a difference between the stochastic training sampler and
the deterministic evaluation sampler.

Directly constraining the policy to preserve the geometry of its original
velocity field may suppress useful task-specific adaptations. Instead,
we reconstruct the flow representation when the measured discrepancy
exceeds a threshold $\gamma$. Endpoint pairs generated by the current RL
policy define the reconstruction targets. This design allows task-driven
adaptation between rectification stages while updating the flow
representation to reflect the policy's current ODE behavior.

\paragraph{Policy-Induced Endpoint Coupling.}
Let $\theta_{\mathrm{ref}}$ denote a frozen snapshot of the current RL policy
parameters, and let $o$ be an observation. We sample a Gaussian initial latent
$X_1\sim\mathcal{N}(0,I)$ and run the policy's deterministic sampler from
flow time $t=1$ to $t=0$. Denote the resulting endpoint by
$X_0=F_{\theta_{\mathrm{ref}}}(o,X_1)$. Repeated sampling defines the empirical, observation-conditioned
coupling
\begin{equation}
  \mathcal{C}_{\theta_{\mathrm{ref}}}
  = \bigl\{(o_i,X_0^i,X_1^i)\bigr\}_{i=1}^{N}.
  \label{eq:policy-induced-coupling}
\end{equation}
For a uniformly
sampled interpolation time $t\sim\mathcal{U}[0,1]$, we form
\begin{equation}
  \widetilde X_t^i=(1-t)X_0^i+tX_1^i,
  \qquad
  u_i^*=\frac{d\widetilde X_t^i}{dt}=X_1^i-X_0^i.
  \label{eq:policy-induced-linear-bridge}
\end{equation}
The target $u_i^*$ is the conditional velocity of the selected linear
bridge, parameterized from the action endpoint to the noise endpoint.
The deterministic policy sampler traverses this pair in the reverse direction.

\paragraph{Action-Subspace Flow Reconstruction.}
Starting from the current policy parameters, we update the same
action-expert velocity field by conditional flow matching on the fixed
policy-induced pairs:
\begin{equation}
  \mathcal{L}_{\mathrm{rect}}(\theta)
  =
  \mathbb{E}_{(o,X_0,X_1)\sim\mathcal{C}_{\theta_{\mathrm{ref}}},\,
  t\sim\mathcal{U}[0,1]}
  \left[
    \frac{\left\|M\odot
    \left(v_\theta(\widetilde X_t,t,o)-(X_1-X_0)\right)\right\|_2^2}
    {\|M\|_1}
  \right],
  \label{eq:policy-induced-rectification-loss}
\end{equation}
where $M$ selects the latent coordinates that contribute to the executed
action.
The full latent is retained as the network input, while coordinates
outside $M$ are excluded from the rectification loss. 

The loss in Eq.~\eqref{eq:policy-induced-rectification-loss} trains the
velocity field to approximate the conditional velocity associated with
the selected linear endpoint coupling. The bridge paths and endpoints
are defined by the current RL policy. Rectification is intended to
improve compatibility between this representation and the linear-path
score approximation used in subsequent SDE-based optimization; its
effect is examined empirically.

\begin{figure}[t]
  \centering
  \begin{tikzpicture}[
    >=Stealth, font=\small,
    process/.style={draw, rounded corners, minimum width=3.15cm,
      minimum height=1.0cm, align=center},
    data/.style={draw, ellipse, minimum width=3.55cm,
      minimum height=1.0cm, align=center},
    flowlabel/.style={font=\scriptsize, align=center, fill=white,
      inner sep=2pt}
  ]
    \node[process] (policy) {Frozen current RL policy\\$v_{\theta_{\mathrm{ref}}}$};
    \node[data, right=3.8cm of policy] (pairs) {Policy-induced pairs\\
      $\mathcal{C}_{\theta_{\mathrm{ref}}}=\{(o_i,X_0^i,X_1^i)\}$};
    \node[process, below=2.1cm of pairs] (rect) {Action-subspace rectification\\
      $v_{\theta_{\mathrm{rec}}}$};
    \node[process, left=3.8cm of rect] (ppo) {PPO updates\\
      next RL stage};
    \draw[->, thick] (policy) -- node[flowlabel, above=3pt]
      {deterministic ODE\\sampling} (pairs);
    \draw[->, thick] (pairs) -- node[flowlabel, right=5pt]
      {linear bridge +\\flow matching} (rect);
    \draw[->, thick] (rect) -- node[flowlabel, below=3pt]
      {initialize next PPO\\stage} (ppo);
    \draw[->, thick] (ppo) -- node[flowlabel, left=5pt]
      {when marginal discrepancy\\exceeds the threshold} (policy);
  \end{tikzpicture}
  \caption{Policy-induced flow rectification between PPO stages, triggered by marginal discrepancy.}
  \label{fig:flow-rectification-loop}
\end{figure}
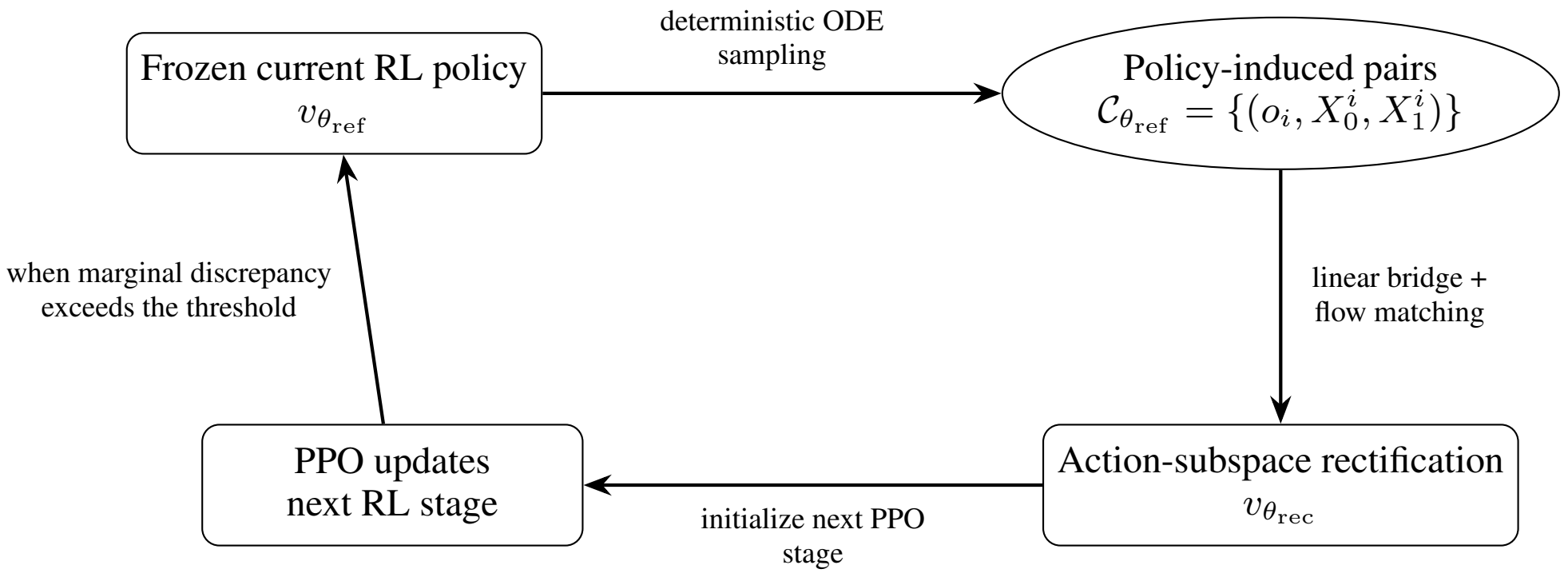

\paragraph{Periodic Rectification During RL.}
We trigger rectification when the measured SDE--ODE marginal
discrepancy exceeds $\gamma$. At each trigger, we collect endpoint pairs
using a frozen snapshot $v_{\theta_{\mathrm{ref}}}$ of the current policy and
update the same action expert, yielding rectified parameters $\theta_{\mathrm{rec}}$.
The updated velocity field is then used for subsequent RL optimization.
This threshold-based schedule avoids reconstruction after every PPO
update while allowing the representation to track accumulated policy
changes. Figure~\ref{fig:flow-rectification-loop} summarizes the procedure.

\subsection{Outcome-Balanced Policy Optimization}

\label{sec:outcome_balanced_ppo}

Standard PPO does not explicitly address two sources of imbalance in
sparse-reward, multi-task manipulation. First, tasks differ in difficulty,
yet the objective does not explicitly allocate additional positive learning
signal to tasks with lower success rates. Second, positive and negative
advantages can differ in magnitude and produce asymmetric policy updates,
as illustrated by the sign-conditioned log-ratio statistics in
Figure~\ref{fig:advantage-pairing}. We address these two sources of
imbalance through task-aware weighting and sign-preserving scaling,
respectively.

\subsubsection{Difficulty-Aware Allocation of Positive Learning Signals}
In multi-task manipulation, a shared policy learns from tasks with
different levels of difficulty. Their contributions to the standard
objective depend on the sampled data and associated advantages. Rare
positive-advantage samples from difficult tasks may consequently receive
limited emphasis relative to the more frequent signals from easier tasks.

\begin{figure}
    \centering
    \includegraphics[width=0.8\linewidth]{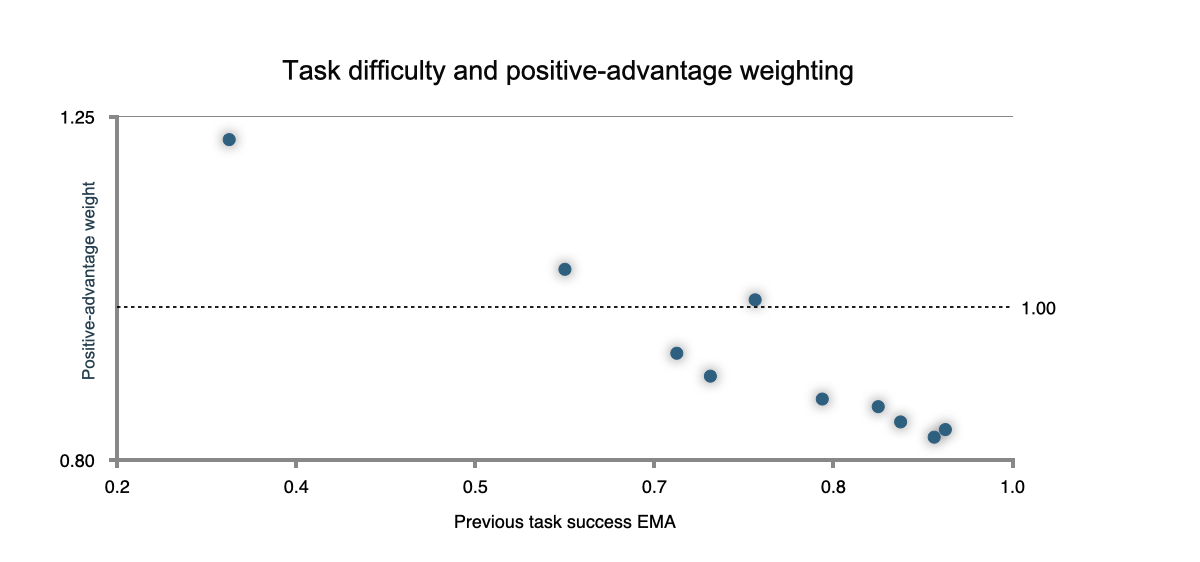}
    \caption{Task-wise EMA success estimates and corresponding positive-advantage weights at a representative training step. Lower success estimates receive larger bounded weights; the plot illustrates the allocation rule rather than an isolated causal effect on success.}
    \label{fig:task-ema-weighting}
\end{figure}

To address this issue, we maintain an exponential moving average (EMA) of the success rate for each task \(k\):
\begin{equation}
    s_k^{(n)} = \beta s_k^{(n-1)} + (1-\beta)\hat{s}_k^{(n)}, \quad s_k^{(0)}=0
\end{equation}
where \(\hat{s}_k^{(n)}\) is the observed success rate of task \(k\) in rollout batch \(n\), and \(\beta\) controls the smoothing horizon. We convert the preceding EMA estimate into a bounded task weight:
\begin{equation}
\tilde{w}_k^{(n)} = \max(1-s_k^{(n-1)}, \epsilon),\qquad
w_k^{(n)} =
\operatorname{clip}\!\left(
\frac{\tilde{w}_k^{(n)}}{\frac{1}{K}\sum_{j=1}^{K}\tilde{w}_j^{(n)}},
w_{\min}, w_{\max}
\right).
\end{equation}
The weights depend on the preceding EMA estimates, preventing the current
rollout from determining its own weights. Figure~\ref{fig:task-ema-weighting}
illustrates the relationship between estimated success rates and
positive-advantage weights. We apply these weights only to positive advantages:
\[
A_{k,i}^{+} \leftarrow w_k^{(n)} A_{k,i}^{+},
\qquad
A_{k,i}^{-} \leftarrow A_{k,i}^{-}.
\]
This weighting reallocates positive learning signal toward tasks with
lower estimated success rates. Negative advantages remain unchanged at
this stage. Bounding \(w_k\) limits the influence of noisy success estimates.

\subsubsection{Sign-Preserving Balancing of Positive and Negative Advantages}

\begin{figure}[t]
    \centering
    \includegraphics[width=0.88\linewidth]{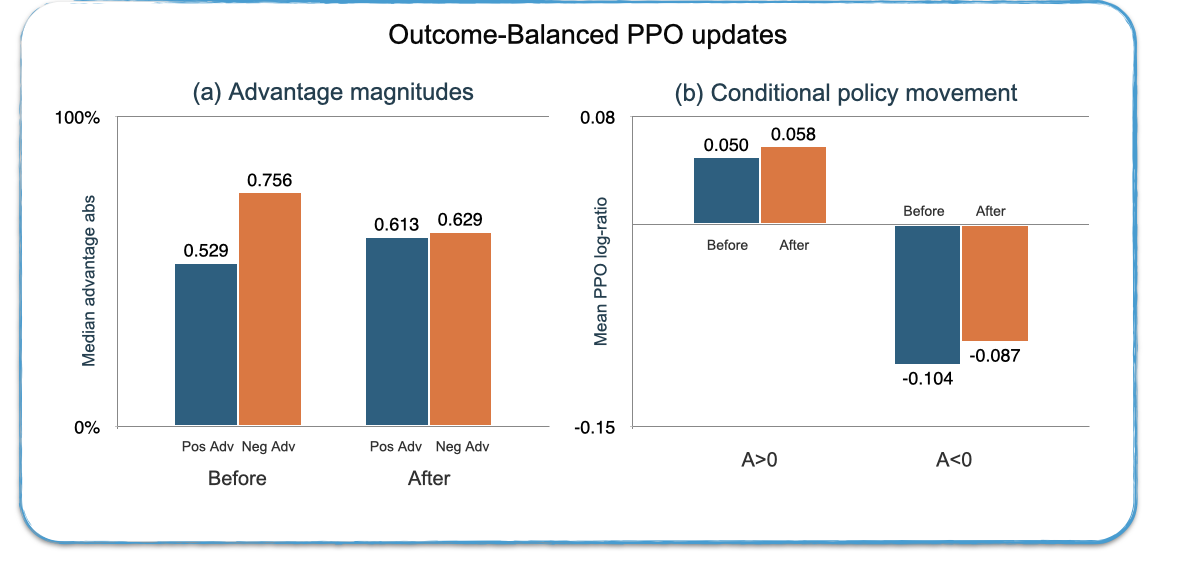}
    \caption{Paired advantage scaling and sign-conditioned PPO log-ratios.
(a) Median magnitudes of positive and negative advantages before and after
paired scaling. (b) Sign-conditioned median log-ratios under standard
and paired advantage normalization.}
    \label{fig:advantage-pairing}
\end{figure}

A second source of imbalance arises within a rollout batch. 
Standard batchwise advantage normalization uses a shared mean and standard deviation:
\[
\hat{A}_i = \frac{A_i-\mu_{\mathcal{B}}}
{\sqrt{\sigma_{\mathcal{B}}^2+\varepsilon}}.
\]
As illustrated in Figure~\ref{fig:advantage-pairing}(a), positive and
negative advantages can have different magnitude scales. Centering all
samples around a shared batch mean does not explicitly match these
scales and may change the sign of a positive advantage below
\(\mu_{\mathcal{B}}\). We therefore introduce sign-preserving paired
advantage normalization.

Let \(\mathcal{B}^{+}=\{i:A_i>0\}\) and \(\mathcal{B}^{-}=\{i:A_i<0\}\) denote the positive- and negative-advantage samples. Their mean absolute magnitudes and the mean absolute advantage across valid samples are
\[
m_{+} = \frac{1}{|\mathcal{B}^{+}|}\sum_{i\in\mathcal{B}^{+}}|A_i|,
\qquad
m_{-} = \frac{1}{|\mathcal{B}^{-}|}\sum_{i\in\mathcal{B}^{-}}|A_i|,
\qquad
m = \frac{1}{|\mathcal{B}|}\sum_{i\in\mathcal{B}}|A_i|.
\]

We transform each group as
\[
\tilde{A}_i =
\begin{cases}
A_i \dfrac{m}{m_{+}}, & A_i>0,\\[6pt]
A_i \dfrac{m}{m_{-}}, & A_i<0,\\[6pt]
0, & A_i=0.
\end{cases}
\]
This transformation preserves the sign of every nonzero advantage and
matches the mean absolute magnitudes of the positive and negative groups.
It is applied after task-specific positive-advantage weighting, retaining
the relative weights among positive samples across tasks. The operation
equalizes group-level advantage scales without enforcing equal sample
counts or equal gradient contributions.

\section{Experiments}

\subsection{Experimental Setup}
\paragraph{Policy Backbone.}

We use $\pi_{0.5}$, which combines a PaliGemma-based VLM with a
flow-matching action expert conditioned on images, language, and
tokenized proprioception~\citep{black2025pi05}.
Figure~\ref{fig:method-overview} summarizes the training architecture.
During RL, SDE rollouts provide samples for updating the action expert
and critic. Deployment uses ODE-based action generation and does not
require the critic.

\paragraph{Simulation Benchmarks.}

We evaluate manipulation policies in four complementary settings.
LIBERO comprises language-conditioned tasks with spatial, object, and
goal variations, as well as extended manipulation
sequences~\citep{liu2023libero}. ManiSkill supports contact-rich
simulation with heterogeneous objects, configurable scenes, and
parallel visual rendering~\citep{tao2024maniskill3}. MetaWorld evaluates
a broad range of motor skills across fifty manipulation task families
with a shared Sawyer embodiment~\citep{yu2020metaworld}. CALVIN tests
the sequential execution of language-specified subtasks in a persistent
workspace~\citep{mees2022calvin}. These settings assess language grounding,
physical interaction, skill diversity, and continuity across subtasks.

\paragraph{Real2Sim2Real Setup.}

We additionally investigate an FR3 carrot-to-plate task using a
Real2Sim2Real pipeline. Physical observations and object measurements
guide the construction of a corresponding ManiSkill/SAPIEN scene.
Simulated demonstrations provide task-specific supervised fine-tuning
(SFT) data, followed by RL in simulation and evaluation of the resulting
checkpoints on the physical robot. Section~\ref{sec:real-world-experiments}
describes the scene, training procedure, deployment, and evaluation.

\paragraph{Hyperparameter Settings.}
For difficulty-aware allocation, we set $\epsilon=10^{-6}$,
$\beta=0.9$, $w_{\min}=0.75$, and $w_{\max}=1.25$. 

\subsection{Simulation Experiments}
\label{sec:simulation_experiments}

We evaluate the proposed framework on LIBERO, ManiSkill, CALVIN,
and MetaWorld. The benchmarks span diverse observations, task
structures, action spaces, and episode horizons. We choose $\pi_{RL}$~\cite{pi_rl} as baseline, which is SOTA in the field of RL for flow-based VLA. We also compare with
$\pi_{0.5}$-SFT and ReFPO~\cite{refpo}. For each benchmark, we evaluate
1,000 episodes per seed across 10 seeds, yielding 10,000 evaluation
episodes in total, and report the average success rate.

\begin{table}[t]
    \centering
    \caption{
        Success rates (\%) across four simulated environments.
        The LIBERO score is the unweighted average of LIBERO-Long,
        LIBERO-Object, LIBERO-Goal, and LIBERO-Spatial.
        The overall average is the unweighted mean across the four
        environment columns. Best results are shown in bold.
    }
    \label{tab:main_simulation_results}
    \small
    \setlength{\tabcolsep}{5pt}
    \renewcommand{\arraystretch}{1.1}
    \begin{tabular}{lccccc}
        \toprule
        Method
        & LIBERO
        & ManiSkill
        & MetaWorld
        & CALVIN
        & Average \\
        \midrule
        $\pi_{0.5}$-SFT
        & 77.1 & 40.1 & 43.8 & 61.3 & 55.6 \\
        \textsc{$\pi_{RL}$}
        & 97.9 & 90.9 & 70.7 & 87.0 & 86.6 \\
        ReFPO & 92.3 & 86.5& 69.5& 85.7  & 83.5\\
        \textbf{Ours}
        & \textbf{98.4}
        & \textbf{95.0}
        & \textbf{73.4}
        & \textbf{89.0}
        & \textbf{88.9} \\
        \midrule
        $\Delta$ vs. $\pi_{RL}$
        & +0.5 & +4.1 & +2.7 & +2.0 & +2.3 \\
        \bottomrule
    \end{tabular}
\end{table}

\paragraph{Comparison with Baseline Policies.}
The four-benchmark average increases from 55.6\% for $\pi_{0.5}$-SFT
to 86.6\% for $\pi_{RL}$ and 88.9\% for our method. The reported
improvements over $\pi_{0.5}$-SFT are 21.3 percentage points on LIBERO,
54.9 on ManiSkill, 29.6 on MetaWorld, and 27.7 on CALVIN.
Relative to $\pi_{RL}$, the corresponding gains are 0.5, 4.1, 2.7,
and 2.0 percentage points. The smaller improvement on LIBERO is
consistent with the baseline success rate of 97.9\%, which leaves
less scope for further gains. These results are aggregated over the
10-seed evaluation protocol described above.

\paragraph{Component Ablations.}
Table~\ref{tab:component_ablations} reports a progressive ablation
of the proposed components. Unlike the four-benchmark average in
Table~\ref{tab:main_simulation_results}, this table assigns equal
weight to seven entries: ManiSkill, MetaWorld, CALVIN, and the four
LIBERO suites. Adding difficulty-aware allocation (DAA) increases
this macro average from 91.4\% to 92.0\%, with the largest individual
gain on ManiSkill (+2.3 percentage points). Adding paired advantage
normalization further increases the average to 92.7\%, with gains
of 1.3, 1.4, and 1.2 percentage points on ManiSkill, MetaWorld, and
CALVIN, respectively, over the preceding configuration.
Figure~\ref{fig:advantage-pairing} illustrates the associated
advantage magnitudes and sign-conditioned policy log-ratios.
Finally, periodic rectification increases the macro average to
93.0\% and improves the point estimates in all four benchmark
families. These results support the utility of the components
within the evaluated training procedure.

\begin{table*}[t]
    \centering
    \caption{
        Success rates (\%) on ManiSkill, MetaWorld, CALVIN, and four LIBERO suites.
        Each method column adds the indicated component to the preceding
        configuration. Macro average is the unweighted mean across the seven
        entries. Gain is the absolute change in percentage points (p.p.)
        of the final configuration over $\pi_{RL}$.
    }
    \label{tab:component_ablations}
    \small
    \setlength{\tabcolsep}{5pt}
    \renewcommand{\arraystretch}{1.15}
    \begin{tabular*}{\textwidth}{@{\extracolsep{\fill}}lccccc}
        \toprule
        Benchmark
        & \shortstack{$\pi_{RL}$\\(Baseline)}
        & \shortstack{+ DAA}
        & \shortstack{+ Paired\\Adv. Norm.}
        & \shortstack{+ Periodic\\Rectification}
        & \shortstack{$\Delta$ vs. $\pi_{RL}$\\(p.p.)} \\
        \midrule
        ManiSkill
        & 90.9 & 93.2 & 94.5 & \textbf{95.0} & +4.1 \\
        MetaWorld
        & 70.7 & 71.4 & 72.8 & \textbf{73.4} & +2.7 \\
        CALVIN
        & 87.0 & 87.4 & 88.6 & \textbf{89.0} & +2.0 \\
        LIBERO-Object
        & \textbf{100.0} & \textbf{100.0} & \textbf{100.0}
        & \textbf{100.0} & +0.0 \\
        LIBERO-Goal
        & 98.8 & 99.0 & \textbf{99.6} & \textbf{99.6} & +0.8 \\
        LIBERO-Spatial
        & 99.6 & 99.6 & \textbf{99.8} & \textbf{99.8} & +0.2 \\
        LIBERO-Long
        & 93.0 & 93.4 & 93.6 & \textbf{94.1} & +1.1 \\
        
        \midrule
        Average
        & 91.4 & 92.0 & 92.7 & \textbf{93.0} & +1.6 \\
        \bottomrule
    \end{tabular*}
\end{table*}

\paragraph{Comparison with Joint Reflow Regularization.}
ReFPO~\citep{refpo} incorporates reflow into FPO~\citep{fpo}, which
does not require explicit evaluation of the policy likelihood ratio.
To compare the reconstruction mechanisms within a shared PPO framework,
we implement a ReFPO-inspired regularizer on top of the $\pi_{RL}$
baseline, using the joint objective
$\mathcal L_{\mathrm{joint}}
=\mathcal L_{\mathrm{baseline}}+\lambda\mathcal L_{\mathrm{rect}}$.
Following the ReFPO setting adopted in this experiment, we use
$\lambda=0.04$. The comparison configuration adds only periodic
rectification to the same baseline.
Table~\ref{tab:two_rec_results} reports higher point estimates for
periodic rectification across all four benchmarks. Under the tested
configurations, the results favor a separate reconstruction stage
over joint reflow regularization.

\begin{table}[htp]
    \centering
    \caption{
        Success rates (\%) for periodic rectification and ReFPO-inspired
        joint reflow regularization added to the same PPO baseline.
        Average is the unweighted mean across the four benchmarks.
        Best results are shown in bold.
    }
    \label{tab:two_rec_results}
    \small
    \setlength{\tabcolsep}{5pt}
    \renewcommand{\arraystretch}{1.1}
    \begin{tabular}{lccccc}
        \toprule
        Method
        & LIBERO
        & ManiSkill
        & MetaWorld
        & CALVIN
        & Average \\
        \midrule
        
        \shortstack[l]{Baseline + Joint Reflow\\(ReFPO-inspired)} 
        & 96.5 & 88.7 & 69.6 & 84.0 & 84.7 \\
        \shortstack[l]{Baseline + Periodic\\Rectification (Ours)}
        & \textbf{98.1}
        & \textbf{92.4}
        & \textbf{71.8}
        & \textbf{87.9}
        & \textbf{87.6} \\
        \bottomrule
    \end{tabular}
\end{table}

\paragraph{Comparison with Adaptive Clipping.}

We compare DAA with paired advantage normalization against DAA
with BAPO-style adaptive clipping~\citep{bapo} to evaluate different
mechanisms for handling positive and negative advantage signals.
Both configurations share the policy initialization, critic
architecture, GAE estimator, DAA weighting rule, and reward function.
Flow rectification is disabled in both configurations to isolate the
advantage-balancing mechanism.

Both methods start from uncentered GAE advantages and apply DAA
before their respective balancing operations. Our configuration
rescales positive and negative advantages to a shared mean absolute
magnitude while preserving their signs and retaining the standard
PPO clipping bounds. The comparison configuration instead retains
the DAA-weighted advantages and adaptively adjusts the clipping
bounds using a BAPO-inspired balancing rule. BAPO was developed
for LLM reinforcement learning; here, we adapt its clipping mechanism
to the Gaussian flow-transition likelihood ratios used by our
PPO implementation. Table~\ref{tab:advantage_balancing_comparison}
reports the resulting success rates across four simulation
benchmarks under a common evaluation protocol.

\begin{table}[t]
    \centering
    \caption{
        Comparison of paired advantage normalization and
        BAPO-style adaptive clipping with shared DAA.
        Flow rectification is disabled in both configurations.
        LIBERO is averaged over its four suites; Average is the
        unweighted mean across the four benchmark columns.
    }
    \label{tab:advantage_balancing_comparison}
    \small
    \setlength{\tabcolsep}{4pt}
    \renewcommand{\arraystretch}{1.15}
    \begin{tabular}{lccccc}
        \toprule
        Method
        & LIBERO
        & ManiSkill
        & MetaWorld
        & CALVIN
        & Average \\
        \midrule
        \shortstack[l]{DAA + BAPO-inspired\\Clipping}
        & 96.6 & 89.2 & 70.8 & 86.8 & 85.9 \\
        \shortstack[l]{DAA + Paired Norm.\\(Ours)}
        & \textbf{98.3} & \textbf{93.6} & \textbf{72.9} & \textbf{88.5}& \textbf{88.3} \\
        \bottomrule
    \end{tabular}
\end{table}
\paragraph{Sensitivity to the Rectification Threshold.}
The threshold $\gamma$ controls the trade-off between the magnitude
of the tolerated SDE--ODE discrepancy and the frequency of
reconstruction. A smaller threshold triggers more frequent
rectification, potentially limiting discrepancy at the cost of
additional computation. A larger threshold reduces the frequency
of reconstruction while permitting greater discrepancy.

We evaluate three threshold values on LIBERO and ManiSkill.
Table~\ref{tab:threshold_chosen} reports success rates and the
mean number of PPO iterations between rectification stages.
Among the tested settings, $\gamma=0.003$ yields the highest
point estimates on both benchmarks and a longer reconstruction
interval than $\gamma=0.001$.

\begin{table}[htp]
    \centering
    \caption{
        Success rates (\%) on LIBERO and ManiSkill and the reported
        mean interval between rectification stages, measured in PPO iterations.
        Best results are shown in bold.
    }
    \label{tab:threshold_chosen}
    \small
    \setlength{\tabcolsep}{5pt}
    \renewcommand{\arraystretch}{1.1}
    \begin{tabular}{lccc}
        \toprule
        Threshold
        & LIBERO
        & ManiSkill
        & \shortstack{Mean Interval\\(PPO Iterations)}
         \\
        \midrule
        
        \textsc{$\gamma=0.001$}
        & 98.2 & 94.9 &  4.6\\
        \textbf{$\gamma=0.003$}
        & \textbf{98.4}
        & \textbf{95.0}
        & 16.6
       \\
        $\gamma=0.006$
        & 97.4
        & 94.0
        & 25.9\\
        \bottomrule
    \end{tabular}
\end{table}

\subsection{Real-World Experiments}
\label{sec:real-world-experiments}

\paragraph{Physical Task and Simulation.}
We evaluate carrot-to-plate transfer using a Franka Research 3 (FR3)
arm with a two-finger gripper. The physical workspace contains a
carrot-shaped object with a body length of approximately 145\,mm and
mass of 50\,g, a shallow red plate with outer and inner diameters of
approximately 156\,mm and 152\,mm, respectively, and a light-yellow
tabletop. A fixed front-facing Orbbec camera provides $640\times480$
RGB observations. We construct the corresponding ManiSkill/SAPIEN
task from workspace photographs and video, a recorded policy-camera
frame, and measured object dimensions~\citep{tao2024maniskill3}.
The scene uses the FR3 articulation, dimension-matched collision
bodies for the carrot and plate, and colored visual proxies.
The carrot leaves contribute only to appearance; contact and
placement checks use the carrot body.

Policy observations are rendered through the ManiSkill-GS interface
used by $\pi_{\mathrm{RL}}$~\citep{pi_rl}. The tabletop is represented
by an analytic support surface comprising 44,744 Gaussians, with
color matched to the physical workspace. We use the physical
camera intrinsics and align the simulated camera pose with the
recorded policy-camera frame. SAPIEN simulates articulation,
contact, and object motion. At reset, the plate remains fixed,
while the carrot position varies within $\pm40\,\mathrm{mm}$
along each tabletop axis, with randomized yaw. The sparse
simulation reward requires a bilateral grasp, a lift of at least
30\,mm, subsequent release and contact with the plate, and a
nearly stationary arm. The placement criterion permits the carrot
body to extend up to 40\,mm beyond the plate's inner radius.

\begin{figure*}
    \centering
    \includegraphics[width=\textwidth]{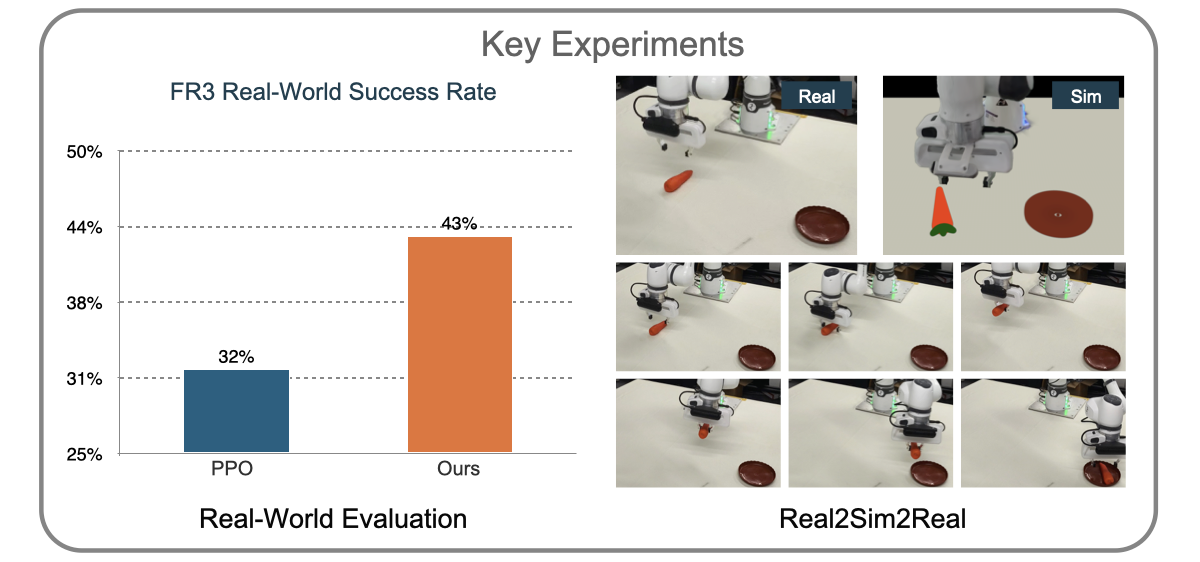}
    \caption{FR3 carrot-to-plate transfer. Left: success rates over 100 physical trials per RL checkpoint. The branch labeled ''Ours'' evaluates a continuation with paired advantage normalization and failed-state replay. The bar-chart vertical axis begins at 25\%. Right: physical and simulated workspaces, followed by representative frames from a real-robot rollout.}
    \label{fig:fr3-real-world}
\end{figure*}
\FloatBarrier

\paragraph{Training and Deployment.}
We collect 20 successful motion-planner trajectories in simulation
and validate them through replay, yielding 3,401 demonstration
frames. Normalization statistics computed from these demonstrations
are shared across checkpoints. Starting from $\pi_{0.5}$, we perform
1,000 full-policy SFT updates. The resulting checkpoint serves as
the no-RL reference and the common initialization for both RL
branches. The standard branch uses RLinf's Flow-SDE PPO with
generalized advantage estimation for 100 epochs. The VLM remains
frozen, while the action expert and critic are optimized. The
modified branch enables paired
advantage normalization and failed-state replay. Paired normalization rescales the advantage signal;
failed-state replay revisits indexed initial conditions associated
with unsuccessful episodes.

During physical deployment, each frozen checkpoint receives the
same language instruction, an Orbbec image, seven joint angles,
and two finger positions. We retain the demonstration-set
normalization and generate five-action chunks using five-step
ODE integration. Each seven-dimensional action specifies
end-effector translation, rotation, and gripper control.
The FR3 client converts the tool-center-point (TCP) target to
the driver end-effector target and executes commands through
Polymetis at a target rate of 10\,Hz. All checkpoints share
the camera, action conversion, executor, and task-reset procedure.

\paragraph{Real-Robot Results.}
We evaluate each RL checkpoint over 100 physical trials using
the same carrot and plate placement range, success criterion,
and deployment mode. A trial is successful if the robot grasps
the carrot, transfers it to the plate, and releases it inside.
Standard PPO succeeds in 32/100 trials (32\%), while the modified
continuation succeeds in 43/100 trials (43\%), corresponding to
an increase of 11 percentage points. The SFT-only policy does
not complete the task during physical execution.
Figure~\ref{fig:fr3-real-world} presents these results alongside
the workspace appearance and a representative manipulation
sequence. This experiment evaluates the combination of paired
advantage normalization and failed-state replay.

\section{Conclusion}

We presented an online reinforcement learning framework for flow-based
vision-language-action policies that combines Outcome-Balanced PPO
with policy-induced flow rectification. Outcome-Balanced PPO addresses
uneven learning signals through task-aware positive-advantage weighting
and sign-preserving advantage scaling. Policy-induced rectification
reconstructs the same action expert from current-policy ODE endpoint
pairs in a separate training stage. The two mechanisms provide
independent control over learning-signal allocation and flow
reconstruction within a unified post-training procedure.

Experiments across diverse simulated manipulation benchmarks support
the utility of the framework and highlight the importance of
considering both the distribution of policy-improvement signals and
the evolution of the flow representation during online optimization.
Future work should characterize when policy-induced reconstruction
improves ODE--SDE sampling consistency, develop reconstruction criteria
that account for computational cost, and evaluate the complete
framework across a broader range of real-world tasks.

\bibliography{iclr2027_conference}
\bibliographystyle{iclr2027_conference}

\end{document}